\documentclass[11pt,a4paper]{article}
\usepackage{times,latexsym}
\usepackage{url}
\usepackage[T1]{fontenc}

\usepackage[acceptedWithA]{tacl2021v1}

\usepackage{xspace,mfirstuc,tabulary}

\newif\iftaclinstructions
\taclinstructionsfalse 
\iftaclinstructions
\renewcommand{\confidential}{}
\renewcommand{\anonsubtext}{(No author info supplied here, for consistency with
TACL-submission anonymization requirements)}
\newcommand{\instr}
\fi

\iftaclpubformat 

\else

\fi

\title{\ourdata: Information-Seeking Conversations \\with Mixed-Initiative Interactions}

\usepackage{microtype}
\usepackage{xspace}
\usepackage{pifont}
\newcommand{\cmark}{\ding{51}}%
\newcommand{\xmark}{\ding{55}}%

\usepackage{tikz}
\newcommand*\halfcirc[1][1ex]{%
  \begin{tikzpicture}
  \draw[fill] (0,0)-- (90:#1) arc (90:270:#1) -- cycle ;
  \draw (0,0) circle (#1);
  \end{tikzpicture}}
  
\usepackage{amsmath,amssymb,amsthm} 
\usepackage{caption}
\usepackage{subcaption}
\usepackage[font=small]{caption}

\usepackage{hyperref}

\usepackage[utf8]{inputenc}
\usepackage{bbm}

\usepackage{esvect}
\usepackage{svg}
\usepackage{comment}

\usepackage{booktabs, multirow} 
\usepackage{soul}
\usepackage{makecell}
\usepackage{threeparttable}
\usepackage{footnote}
\makesavenoteenv{tabular}

\usepackage{tablefootnote}

\newcommand{\sewon}[1]{{\color{violet}{[{\bf sewon}: #1]}}}

\newcommand{\ourdata}{\textsc{InSCIt}\xspace}

\newcommand\ai{$^{\diamondsuit}$}
\newcommand\uw{$^\spadesuit$}
\newcommand\ms{$^\clubsuit$}
\newcommand\aspace{\hspace{.75em}}

\author{
    Zeqiu Wu \uw\aspace
    Ryu Parish \uw\aspace
    Hao Cheng \ms\aspace
    Sewon Min \uw\aspace \\
    {\bf Prithviraj Ammanabrolu} \ai\aspace
    {\bf Mari Ostendorf} \uw\aspace
    {\bf Hannaneh Hajishirzi} \uw\ai\aspace \\
    \uw University of Washington \aspace
    \ms Microsoft Research \aspace
    \ai Allen Institute for AI\\
    {\tt \{zeqiuwu1,rparish,sewon,ostendor,hannaneh\}@uw.edu} \\
    {\tt chehao@microsoft.com raja@allenai.org}
}

\date{}

\begin{document}
\maketitle
\begin{abstract}
In an information-seeking conversation, a user may ask questions that are under-specified or unanswerable. An ideal agent would interact by initiating different response types according to the available knowledge sources. However, most current studies either fail to or artificially incorporate such agent-side initiative. This work presents \ourdata, a dataset for \textbf{In}formation-\textbf{S}eeking \textbf{C}onversations with mixed-initiative \textbf{I}n\textbf{t}eractions. It contains 4.7K user-agent turns from 805 human-human conversations where the agent searches over Wikipedia and either directly answers, asks for clarification, or provides relevant information to address user queries. The data supports two subtasks, evidence passage identification and response generation, as well as a human evaluation protocol to assess model performance. We report results of two systems based on state-of-the-art models of conversational knowledge identification and open-domain question answering. Both systems significantly underperform humans, suggesting ample room for improvement in future studies.\footnote{We open-source all data and code at \url{https://github.com/ellenmellon/INSCIT}.
}
\end{abstract}

\section{Introduction}

\begin{figure}
\centering
\includegraphics[width=1.0\linewidth]{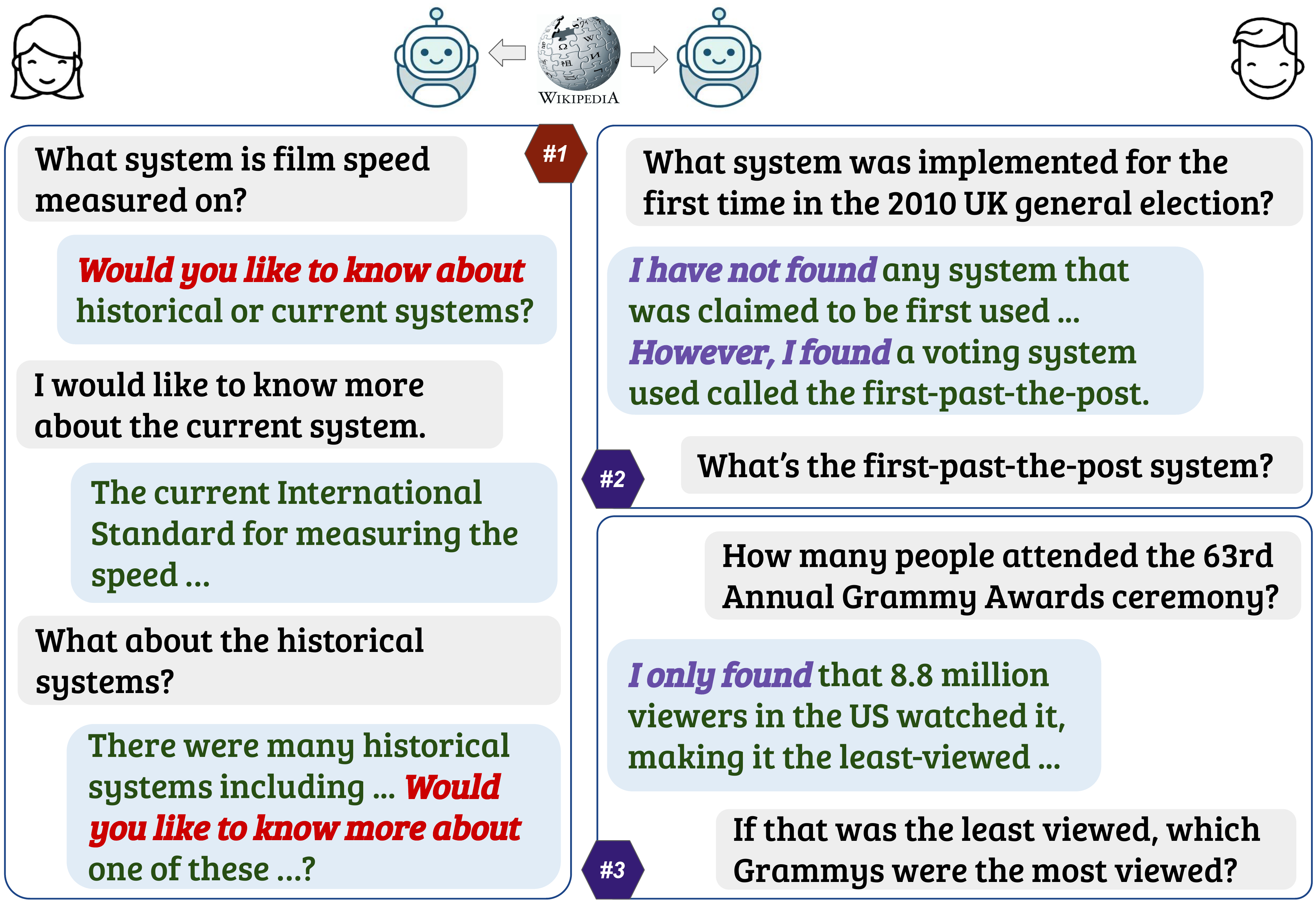}
\caption{\label{fig:teaser} \ourdata examples show that user queries can often be under-specified and require clarification (left), or have no direct answer but where providing relevant information may fulfill users' information needs (right).
}
\end{figure}

Recently, there is increasing interest in developing conversational information-seeking systems \citep{choi-etal-2018-quac, adlakha2022topiocqa, saeidi-etal-2018-interpretation, feng-etal-2020-doc2dial} that assist users in finding information from knowledge sources (e.g., text corpus) via multi-turn conversational interactions.
One important advantage of such conversational information-seeking systems is that users do not need to come up with a very descriptive query 
by themselves \citep{webb_webber_2009, rieser_lemon_2009, Konstantinova-2013-iqa}.
In realistic settings, as shown in \autoref{fig:teaser}, users can start with a request that is under-specified or has no direct answer, and through conversational interactions, the agent can collaboratively guide users to refine (left) or relax their queries and proactively suggest relevant information that may partially satisfy the user's information needs (right).
This collaboration requires a mixed-initiative dialogue, where both the user and agent can direct the flow of the conversation. 

Handling such realistic user requests poses challenges to a conversational agent system.  
A comprehensive search can result in \textit{multiple passages} from the knowledge source, which may provide different components of an answer (possibly incomplete) or multiple answers that surface ambiguities in the user query.
Depending on the available information, the agent needs to use \textit{different strategies}, which might involve summarizing the results, providing partial information, or trying to clarify an ambiguity.
However, existing information-seeking conversation datasets rarely contain conversations where agents initiate different interaction strategies. 
As a result, most conversational question answering (CQA) work focuses on user-initiative interactions, 
where the agent simply responds to user questions with direct answers or uses \texttt{no answer} for out-of-scope queries \citep{choi-etal-2018-quac, reddy-etal-2019-coqa, adlakha2022topiocqa}.
Other work studies clarification questions using artificially-created data, failing to capture natural information-seeking interactions \citep{saeidi-etal-2018-interpretation, feng-etal-2020-doc2dial, aliannejadi-etal-2021-building, Guo2021AbgCoQACA}.
In addition, most of them only rely on a single evidence string or passage for agent response construction.



To support research in mixed-initiative conversations, we introduce \ourdata (pronounced Insight), a dataset for \textbf{In}formation-\textbf{S}eeking \textbf{C}onversations with mixed-initiative \textbf{I}n\textbf{t}eractions, where agents take various strategies, such as providing direct answers (72\%), raising clarifications (13\%), and presenting relevant partial information (13\%), to address users' information needs. 
It contains 805 natural human-human conversations with 4.7K user-agent turns over diverse topics, collected through a scalable annotation pipeline and careful quality control.
To simulate realistic information-seeking scenarios, users write queries with minimal restriction, and human agents decide on different strategies to respond, after searching over the knowledge source (i.e., Wikipedia) for evidence passages.

We formulate two tasks for the conversational agent system: (1) identify a set of evidence passages from Wikipedia, and (2) generate a response grounded in the evidence.
Since handling queries with multiple evidence passages or no direct answer can be open-ended, we emphasize the need for human evaluation, and propose a systematic human evaluation protocol that considers diverse aspects including coherence, factual consistency and information comprehensiveness.

We present two strong baselines based on the state-of-the-art in open-domain question answering \citep{karpukhin-etal-2020-dense, izacard-grave-2021-leveraging} and conversational knowledge identification \citep{wu-etal-2021-dialki}. 
While the systems achieve substantial improvements over a trivial baseline, there is still significant room for improvements, especially for scenarios requiring agent strategies other than providing a direct answer.
Our analysis suggests that the key remaining challenges are improving passage identification and fusing comprehensive information from \textit{multiple passages} by leveraging \textit{different strategies}. We present detailed discussion and avenues for future work.

\section{Related Work}
\label{sec:related}

\paragraph{Information-Seeking Conversations}
The aim of information-seeking conversations is to address the user’s initial and follow-up information needs with grounding in knowledge sources. Table~\ref{tab:data_comparison} compares \ourdata with previous information-seeking conversation datasets. Early CQA work, including QuAC \cite{choi-etal-2018-quac} and CoQA \cite{reddy-etal-2019-coqa}, requires the agent to answer each user question by reading a short passage. DoQA \citep{campos-etal-2020-doqa}, QReCC \citep{anantha-etal-2021-open} and TopioCQA \citep{adlakha2022topiocqa} extend the task to an open-domain setting where the knowledge source is a large document corpus. These studies only consider limited scenarios where the agent provides a direct answer based on a short text span in a single passage, or outputs \texttt{no answer} if there is no direct answer.

\begin{table}
\centering
\small
\resizebox{\columnwidth}{!}{
\begin{tabular}{lc|cc|cccc}\toprule
\multirow{2}{*}{\textbf{Dataset}} & \multirow{2}{*}{\textbf{IR}} &\multicolumn{2}{c}{\textbf{Response Strategy}} & \textbf{H-H} & \textbf{Multi-}  \\
& &\textsc{clar} &\textsc{rel} & \textbf{Dialogue} & \textbf{Evidence}  \\\midrule
\textbf{\ourdata (ours)} & \cmark & \cmark & \cmark & \cmark & \cmark   \\\midrule
QuAC & \xmark & \xmark & \xmark & \cmark & \xmark \\
CoQA & \xmark & \xmark & \xmark & \cmark & \xmark \\
DoQA & \cmark & \xmark & \xmark & \cmark & \xmark \\ 
QReCC & \cmark & \xmark & \xmark & \halfcirc & \xmark \\
TopioCQA & \cmark & \xmark & \xmark & \cmark & \xmark \\
Qulac & \xmark & \cmark & \xmark & \xmark & \xmark \\
ShARC & \xmark & \cmark & \xmark & \xmark & \xmark \\
MultiDoc2Dial & \cmark & \cmark & \xmark & \xmark & \xmark \\
Abg-CoQA & \xmark & \cmark & \xmark & \xmark & \cmark  \\
\bottomrule
\end{tabular}
}
\caption{Comparison of \ourdata with existing datasets of information-seeking conversations.
\textit{IR}, \textit{CLAR}, \textit{REL}, \textit{H-H} stand for \textit{Retrieval Needed}, \textit{Clarification}, \textit{No Direct but Relevant Answer} and \textit{Human-Human}. \halfcirc \xspace indicates the property only applies to part of the dataset.}
\label{tab:data_comparison}
\end{table}

Ambiguous user queries have been observed in single-turn question answering tasks \citep{min2020ambigqa, zhang-choi-2021-situatedqa, sun-etal-2022-conditionalqa}, but these are usually addressed by training a model to predict multiple conditional answers without further interaction. 
A few other studies create artificial conversations to address ambiguous user questions.
For instance, Qulac \citep{Aliannejadi_2019} and the data collected in follow-up work \citep{aliannejadi-etal-2021-building} are based on user queries containing a set of pre-specified multi-faceted entities,
where agents choose from a fixed set of clarification questions that cover these ambiguities.
ShARC \cite{saeidi-etal-2018-interpretation}, Doc2Dial \cite{feng-etal-2020-doc2dial} and MultiDoc2Dial \citep{feng-etal-2021-multidoc2dial} are rule-based information-seeking conversations in the social welfare domain that incorporate agent-side clarifications.
\citet{Guo2021AbgCoQACA} create Abg-CoQA by rewriting conversations in the CoQA dataset to intentionally include ambiguous questions.
In contrast, \ourdata consists of human-human conversations with natural information-seeking user requests and mixed agent initiative to address them.

\citet{penha2019mantis} crawl conversations from Stack Exchange\footnote{\url{https://stackexchange.com/}} that are mixed with information-seeking utterances and casual talk. One grounding document is heuristically obtained for each conversation. In contrast, \ourdata contains validated grounding passages and only goal-oriented agent interactions. 

\paragraph{Knowledge-Grounded Social Chat}
Instead of seeking for information, the user intent in social chat is mostly to conduct casual talk.
Knowledge-grounded social chat systems \citep{Ghazvininejad2018AKN, dinan2018wizard, zhou2018dataset, Moghe2018TowardsEB} incorporate external knowledge with the purpose of making the conversations more engaging and informative. \citet{rodriguez-etal-2020-information} trains a conversational agent to select knowledge to present based on the user's background, in order to maintain the user’s interest in the conversation.

\section{Task Formulations}
\label{sec:tasks}
We define two task formulations for \ourdata, namely \textit{passage identification} and \textit{response generation}. These two tasks mimic how an agent responds to each information-seeking user request, by first searching for relevant information over the knowledge source and then constructing the response based on the gathered information.
Comparing with prior studies on open-domain information-seeking conversations~\citep{anantha-etal-2021-open, adlakha2022topiocqa}, the key challenges in our tasks come from identifying and fusing comprehensive information from \textit{multiple passages} to construct responses using \textit{different strategies}, rather than a single passage and a short answer.

At the $n^{th}$ agent turn, both tasks have the same input: all previous utterances (i.e., dialogue context) $X=[u_1, a_1, u_2, a_2, ..., u_n]$, the corpus of all passage candidates $\mathcal{C}$, and the previously used passages $\{\mathcal{P}_1, \mathcal{P}_2, ..., \mathcal{P}_{n-1}\}$ where each $\mathcal{P}_i = \{p_i^1, p_i^2, ..., p_i^{|P_i|}\}$ is the set of passages used in the $i^{th}$ agent turn $a_i$. 
$\mathcal{C}$ is defined as all textual paragraphs 
(i.e., passages)
in a full Wikipedia dump.\footnote{We use the dump of 04/20/2022.}

For \textit{passage identification}, we require the model to predict a \textit{set} of passages $\Bar{\mathcal{P}}_n$ from $\mathcal{C}$, containing comprehensive and relevant information to the current user request $u_n$ in the dialogue context $X$, which serves as evidence for the \textit{response generation} task---generating the next agent response $\Bar{a}_n$. This is different from the passage retrieval task where only a ranked list of relevant passages is predicted. 
Identifying specific knowledge to be used in the response can be important for model interpretability purposes as well as for evaluating how well a model grounds the response generation in the knowledge source.
Ideally, all factual information contained in $\Bar{a}_n$ should be consistent with $\Bar{\mathcal{P}}_n$, and every passage in $\Bar{\mathcal{P}}_n$ should provide at least one unique information piece as evidence for $\Bar{a}_n$.

In interactive dialogues, 
each predicted evidence $\Bar{\mathcal{P}}_i$ and response $\bar{a}_i$ are used in the dialogue context for later conversations.
However, to use pre-collected dialogues with automatic evaluation metrics, the input context must be the same as that leading to the human reference response. 
This is also consistent with setups in previous information-seeking dialogue studies that are discussed in \S~\ref{sec:related}.
Therefore, the gold $\{\mathcal{P}_i \}$ and $\{a_i\}$ are used here as inputs in testing.

\begin{figure*}
\centering
\includegraphics[width=1.0\linewidth]{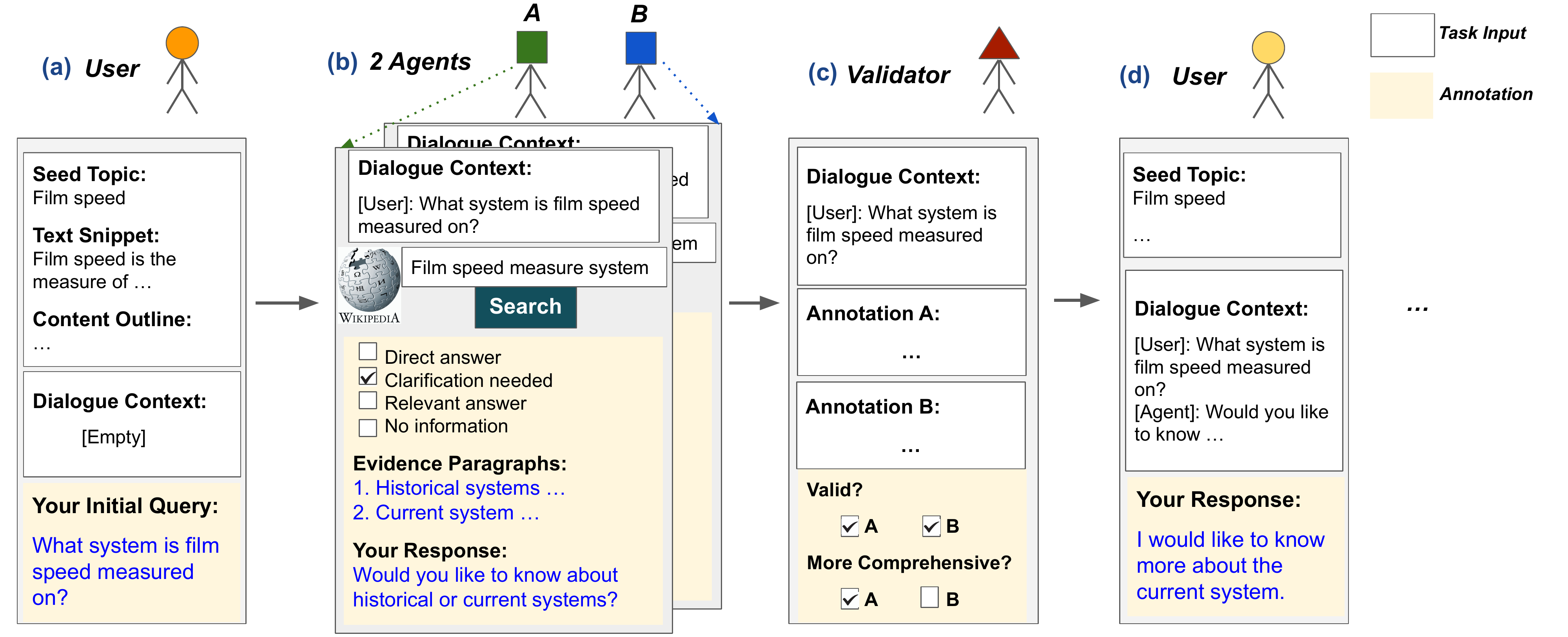}
\caption{\label{fig:annotation} \textcolor{black}{Each conversation is annotated in a series of user $\rightarrow$ agent $\rightarrow$ validator tasks. One worker is dedicated to each user/validator task but two workers work in parallel on the agent turn annotation (see discussion in \S~\ref{sec:annotation}).} }
\end{figure*}

\section{Our Data: \ourdata} 
\label{sec:data}

We introduce \ourdata, a new information-seeking conversation dataset where the agent interprets the user intent and provides comprehensive information grounded in Wikipedia via natural human-human interactions. In this section, we present our data collection pipeline, quality control mechanisms, and analyses that show the characteristics and diversity of the user and agent turns.

\subsection{Data Collection Pipeline}
\label{sec:annotation}
We recruit user, agent and validation workers\footnote{We use Amazon Mechanical Turk (\url{https://www.mturk.com/}) for data collection.} to create and annotate user/agent turns and validate agent annotations, respectively. 
Due to the asymmetric time spent by user and the agent workers in a conversation, 
we design a separate annotation task for each user or agent turn, following \citet{wen-etal-2017-network} to annotate each dialogue in a pipelined fashion.
This framework has proved to be efficient while maintaining the conversation coherence by requiring each worker to read all previous utterances. 
\textcolor{black}{Our data collection has IRB approval and is deemed exempt. }

Figure~\ref{fig:annotation} illustrates the data collection and annotation pipeline. 
Each conversation starts with an initial user turn, where the worker asks a question after reading a text snippet from a seed document.
Then, two agents independently search for relevant passages in Wikipedia, provide a response, and categorize their response.
Validation follows after each user-agent turn.
We refer to the retrieved passages, contributed text, and validations collectively as ``annotations.''
The user/agent/validation process is repeated for 7 turns or until responses are found to be invalid.
Details for each step follow.

\begin{figure*}
\centering
\fontsize{9pt}{9pt}\selectfont
\begin{subfigure}[b]{0.28\linewidth}
\centering
\includegraphics[width=1.0\linewidth]{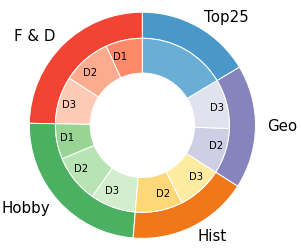}
\end{subfigure}
\hfill
\begin{subfigure}[b]{0.35\linewidth}
\centering
\includegraphics[width=1.0\linewidth]{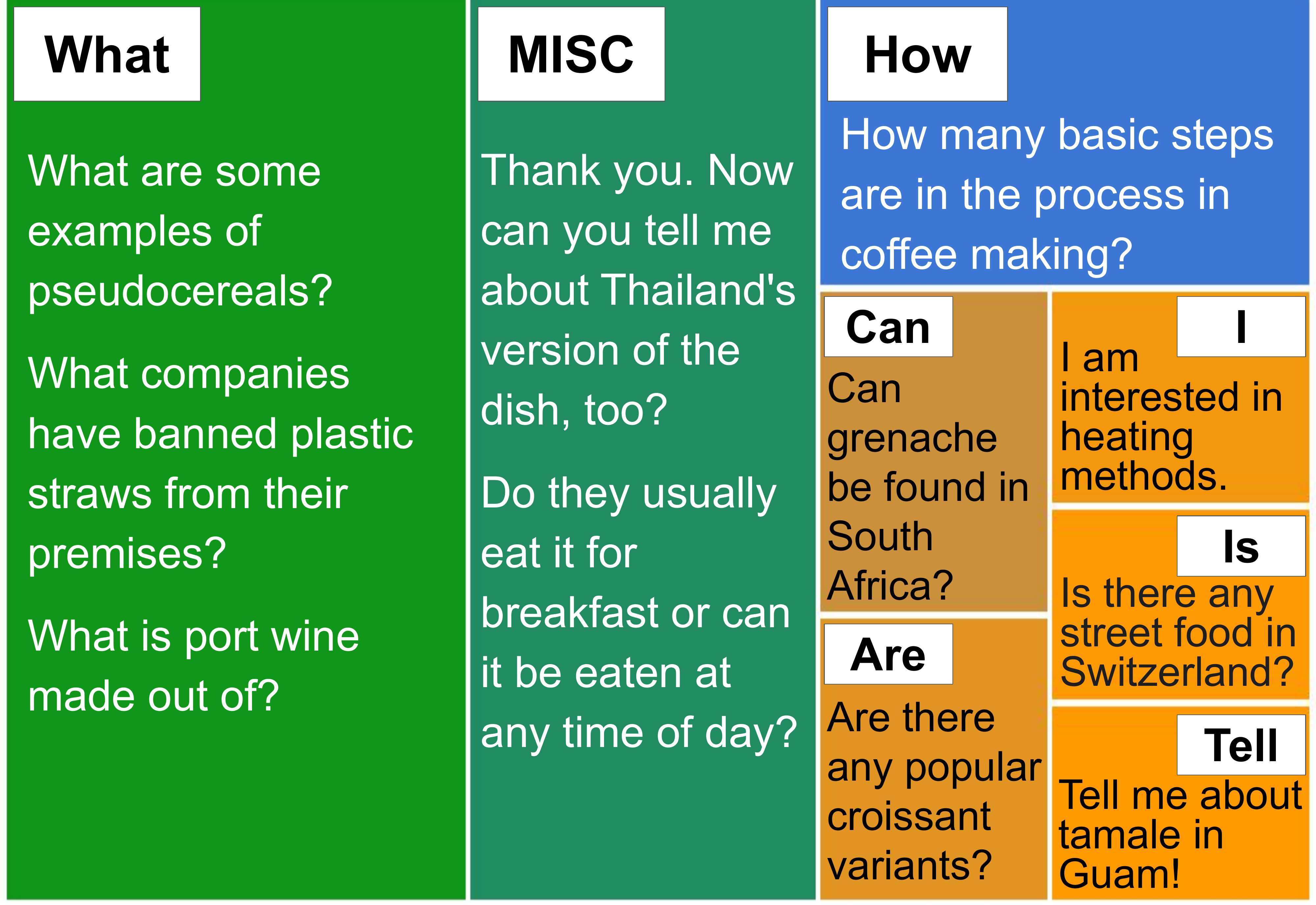}
\end{subfigure}
\hfill
\begin{subfigure}[b]{0.35\linewidth}
\centering
\includegraphics[width=1.0\linewidth]{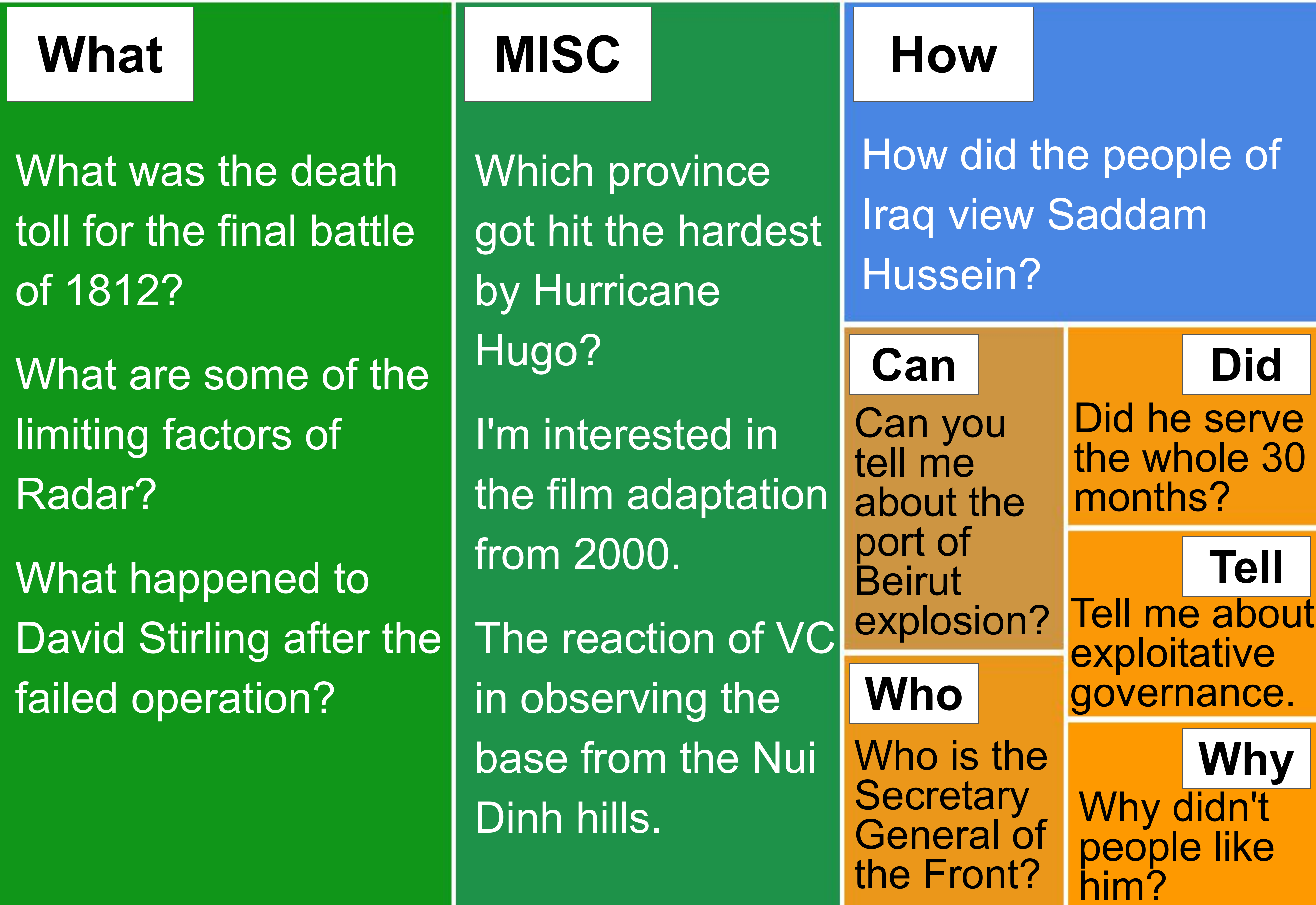}
\end{subfigure}
\caption{\label{fig:data_analysis} Left: seed document topic category breakdown (\textit{D} $\rightarrow$ category \textit{depth}). Middle and right: treemaps of top 7 (and other: MISC) first tokens in user turns from conversations under ``food \& drink'' and ``historical events'' topic categories. \textit{For each figure, the size of each colored area is proportional to its percentage in the data.}}
\end{figure*}

\paragraph{Seed Document Selection}
To diversify conversation topics, we sample seed Wikipedia articles, used for triggering initial user requests, from 5 different topic categories---food and drink, hobby, historical events, geography and weekly top-25 pages. Additionally, we leverage the top-down tree structure of Wikipedia categories\footnote{\url{https://en.wikipedia.org/wiki/Wikipedia:Contents/Categories}} and sample articles at various tree depths under each of the first 4 categories.
Weekly top-25 pages are from Wikipedia weekly reports of 2021.\footnote{\url{https://en.wikipedia.org/wiki/Category:Wikipedia_Top_25_Report}} 
Figure~\ref{fig:data_analysis} (left) shows the distribution of sampled seed documents under each category and their corresponding depths.

\paragraph{User Turn}
Here, a user worker is asked to write an \textit{initial query} or \textit{follow-up response} to continue the existing conversation.
To trigger each conversation (Figure~\ref{fig:annotation} (a)), the user worker is presented with the leading paragraph of a seed article, and is instructed to ask a question they are interested in but cannot find the answer from the paragraph.
The article content outline containing all section titles is also provided to help with the question construction.
The annotation for each following user turn (d) starts after the completion of the previous agent annotation (b) and the validation step (c), based on all previous conversation utterances.

\paragraph{Agent Turn}
Different from the user worker, in addition to the dialogue context,
each agent worker (Figure~\ref{fig:annotation} (b)) is given all evidence paragraphs used by each previous agent turn as additional context.
Then, the worker is told to use the provided search engine\footnote{based on Google Search API from \url{https://developers.google.com/custom-search} and restricted to the \url{https://en.wikipedia.org/} domain.} to find answer(s) from Wikipedia for the current user request. 
They are asked to select all (up to 4) evidence paragraphs from Wikipedia, which they then use to construct their response. 
They are also asked to categorize their response, choosing
one of four response strategies: \{\textit{direct answer} (\textsc{direct}), \textit{clarification} (\textsc{clar}), \textit{relevant answer} (\textsc{rel}), and \textit{no information} (\textsc{ni})\}.
In contrast to a direct and complete answer, we consider a response as a \textit{relevant answer} when the agent finds information that only partially satisfies the user need (e.g., relax a constraint in the request).
For each agent turn, we collect two different annotations to increase reference diversity. 

\paragraph{Validation} After each user turn, we send the two agent annotations to a validator (Figure~\ref{fig:annotation} (c)). 
For each agent turn annotation, the validator  determines whether
i) each selected evidence paragraph is properly used in the response;
ii) the response is factually consistent with the evidence;
iii) the response is coherent to the dialogue context; and
vi) the labeled response strategy is faithfully reflected in the response.
If both are valid, the validator is asked to rate which one is more comprehensive, where a tie is permitted.
An agent response is considered as more comprehensive if it contains more information relevant to the user request.
The more comprehensive (or the only valid) annotation\footnote{We randomly select one if there is a tie.} is then used to continue the conversation.
\textcolor{black}{The annotation is terminated if both annotations are invalid, and we include the conversation up to the previous turn in our data.}

\subsection{Quality Control}
\label{sec:quality}

\paragraph{Worker Qualification}
To recruit agent workers, we manually review > 150 submissions of a qualification task and select 24 highly qualified workers who consistently produce valid annotations during the qualification. \textcolor{black}{The qualification task consists of 12 agent annotation tasks, where each dialogue context is written by the first two authors of this paper. Similarly, we create different qualification tasks to select 35 qualified users and 10 validators who consistently produce reasonable user responses or validations based on our manual review.}

\paragraph{Annotation Control}
To discourage users from chit-chatting or raising inappropriate requests (e.g., too subjective), each agent worker can decide to either continue the conversation or flag their previous user turn as incoherent or an invalid request.
The validation process ensures that only valid
agent annotations are included in our final dataset. 
%
To encourage extensive search for comprehensive information, we assign a bonus to an agent worker if their annotation is labeled as equally or more comprehensive than the other worker. 

We constantly monitor the annotation process and send feedback to workers. 
Our user and agent workers have over 99\% and 96\% average passing validation rate respectively. 
About 13\% of agent annotations are marked as less comprehensive.


\begin{table}
\centering
\normalsize
\resizebox{0.38\textwidth}{!}{
\begin{tabular}{l   c c c c }
\toprule
 & \textbf{Train} & \textbf{Dev} & \textbf{Test} & \textbf{Total}  \\
\midrule
\# Convs & 250 & 86 & 469 & 805  \\
\# Turns & 1443 & 502 & 2767 & 4712  \\
\# Turns / Conv & 5.8 & 5.8 & 5.9 & 5.9 \\
\# References / Turn & 1.8 & 1.6  & 1.6  & 1.7 \\
\# Tokens / User & 10.6 & 10.5 & 10.7 & 10.6 \\
\# Tokens / Agent & 35.7 & 44.3 & 45.1 & 41.9 \\
\# Passages / Agent & 1.5 & 1.7 & 1.6 & 1.6 \\
\bottomrule
\end{tabular}
}
\caption{\label{tab:data} Overall statistics of \ourdata.}
\end{table}

\begin{table*}[h]
\centering
\normalsize
\resizebox{1.0\textwidth}{!}{
\begin{tabular}{p{4cm}  p{16cm}}
\toprule
\textbf{Clarification (\textsc{clar})} \\\midrule
Too long / many answers \newline (86\%) &\dots \textbf{<Agent>:} In the Battle of New Orleans, \dots the Americans had 13 dead, totaling 304 dead. \newline \textbf{<User>:} Were there any long-term consequences that came as a result of the War of 1812? \newline \textbf{<Agent>:} There were many! Would you like to know what they were for \textit{Bermuda, \dots, or Great Britain}? \\\midrule
Ambiguous entity (13\%) &\textbf{<User>:} Washington University is classified as what for its high research activities? \newline \textbf{<Agent>:} Do you want to know about \textit{Washington University in St. Louis or in Baltimore, Maryland}? \\\midrule\midrule
\multicolumn{2}{l}{\textbf{No Direct but Relevant Answer (\textsc{rei})}} \\\midrule
\multirow{2}{*}{\shortstack[l]{Constraint relaxation / \\ No definite answer (70\%)}}&\textbf{<User>:} Was the Matrix franchise adapted from any other work? \newline \textbf{<Agent>:} While not explicitly adapted from another work, \textit{the Matrix did draw heavily on Jean} \dots \\\cmidrule{2-2}
&\dots \textbf{<User>:} Who authored the Torah? \newline \textbf{<Agent>:} \textit{It was originally thought that a majority of the Torah was written by \dots However, it's now thought that \dots though the number of authors is up for debate.} \\ \midrule
Relevant but side info \newline only  (29\%) &\dots \textbf{<User>:} What countries have an ecological footprint under 5 hectares per person? \newline \textbf{<Agent>:} \textit{The world-average ecological footprint in 2013 was 2.8 global hectares per person \dots} But I don't have a list of countries with an ecological footprint under 5 global hectares per person. \\
\bottomrule
\end{tabular}
}
\caption{\label{tab:agent_examples} Examples of clarification and no-direct-but-relevant-answer agent responses. \textit{Factual information from evidence passages is italicized in agent responses.}}
\end{table*}

\begin{table}
\centering
\small
\resizebox{0.35\textwidth}{!}{
\begin{tabular}{l   c c c c }
\toprule
 &  \textsc{direct} & \textsc{clar} & \textsc{rel} & \textsc{ni}  \\
\midrule
\% Turns & 71.5 & 12.7 & 13.1 & 2.7 \\
\# Tokens & 43.7 & 33.5 & 46.6 & 10.6 \\
\# Passages & 1.5 & 2.8 & 1.4 & 0.0 \\
\bottomrule
\end{tabular}
}
\caption{\label{tab:data_by_rtype} Agent response strategy statistics.
 \textsc{direct}, \textsc{clar}, \textsc{rel}, and \textsc{ni} indicate direct answer, clarification, no direct but relevant answer, and no information.
}
\end{table}

\textcolor{black}{
\paragraph{Worker Payment Structure}
We actively communicate with workers throughout the annotation process to clarify any questions they have and to give them feedback. We also check in with them early on to make sure they are satisfied with the pay and bonus structure. Most workers report that they are paid with an hourly rate of 15-20 USD, depending on their annotation speed. We pay 0.2/0.5/0.5 USD for each user/agent/validator annotation, plus a 0.1 USD bonus for each agent annotation if the worker passes validation over 80\% of the time (all qualified).
In addition, we assign a bonus of 0.3 USD to the agent annotation that is marked as equally comprehensive as its peer annotation by the validator, or 0.5 to those marked as more comprehensive or with multiple evidence passages found.\footnote{At the beginning of our training set collection (before the collection of dev/test sets), we only assign a 0.3 USD bonus to agent annotations marked as more comprehensive. After communicating with our workers, we adjust our bonus structure, which leads to more comprehensive agent responses.} On average, we pay over 0.9 USD to each agent annotation. }

\subsection{Data Analysis}
\label{sec:analysis}

We collect 805 conversations, which includes 4712 user-agent turns 
after dropping agent annotations if their evidence passages cannot be found in the post-processed Wikipedia corpus.\footnote{We use wikiextractor to process Wikipedia articles: \url{https://github.com/attardi/wikiextractor}.}
Table~\ref{tab:data} shows summary statistics of the train/dev/test subsets of \ourdata.
Word token counts are based on the \textcolor{black}{spaCy \citep{Honnibal_spaCy_Industrial-strength_Natural_2020}} tokenizer. 
The \textit{test set} contains conversations triggered with seed documents from all 5 topic categories, while the \textit{training} and \textit{dev sets} only contain those from ``food and drink'', ``hobby'' and ``top-25''. 
In the training set, we keep all valid agent annotations as well as their comprehensiveness comparison results. 
In the dev and test sets, we did not include agent responses flagged as less comprehensive during validation.
In addition, as discussed in \S~\ref{sec:quality}, we adjust the worker incentives to obtain more comprehensive responses when collecting dev/test sets, leading to the difference in the average agent turn length.

\subsubsection{Diversity of User and Agent Turns}
\label{sec:data_diversity}
\paragraph{User Request}
We analyze the distribution of wh-words of user questions, as well as non-question user utterances (e.g., responses to clarification). The treemaps in Figure~\ref{fig:data_analysis} (middle and right) show the 7 most frequent leading unigrams of user utterances in ``food \& drink'' and ``historical events'' conversations respectively. ``MISC'' refers to utterances with less frequent leading unigrams.
Each box size is proportional to its percentage in the data.
As shown, most user requests are ``what'' and ``how'' questions.
There are also many user turns starting with words like ``can'' and ``tell'', most of which are responses to agent clarification questions.
The user utterances are fairly long-tailed as ``MISC'' shares a large portion (about 30\%) for both treemaps.
Instead of being mostly factoid questions, 
open-ended user requests are well represented in
\ourdata.

\begin{table*}
\centering
\normalsize
\resizebox{1.0\textwidth}{!}{
\begin{tabular}{p{20.5cm}}
\toprule
\textit{Different Evidence; Different Response Strategies} \\\midrule
\textbf{Dialogue Context:} \dots \textbf{<User>:} What are some of the health consequences of oats? \\\midrule
\textbf{<Agent> (\textsc{clar}):} Oat related health concerns include oat allergies, and gluten-related disorders. Which type of health issue would you like to learn more of? \textit{(evidence from Wikipedia page: Oat Sensitivity)} \\ 
\textbf{<Agent> (\textsc{direct}):} Oat toxicity in people with gluten-related disorders depends on the oat cultivar consumed because the immunoreactivities of toxic prolamins are different among oat varieties. Furthermore, oats are frequently cross-contaminated with the other gluten-containing cereals \dots \textit{(evidence from Wikipedia page: Oat)} \\\bottomrule \addlinespace \toprule
\textit{Same Evidence; Different Response Strategies} \\\midrule
\textbf{Dialogue Context:} \dots \textbf{<User>:} Who is the major competitor of PepsiCo? \\ \midrule
\textbf{<Agent> (\textsc{clar}):} Would you like to know PepsiCo's major competitor in the beverage market or the snack food market?  \\
\textbf{<Agent> (\textsc{direct}):} PepsiCo has two main competitors in different market sectors. The Coca-Cola Company is its largest competitor in the beverage market, and Kraft Foods is its largest competitor in the snack foods market. \\\midrule\midrule
\textbf{Dialogue Context:} \dots \textbf{<User>:} Were the directors involved in any controversy? \\\midrule
\textbf{<Agent> (\textsc{rel}):} I couldn't find anything about themselves in any controversy but their film \dots had a controversial storyline and themes. \\ 
\textbf{<Agent> (\textsc{direct}):} The Wachowskis' V for Vendetta was considered to have controversial storyline and themes, but have been both criticized and praised by sociopolitical groups. \\
\bottomrule
\end{tabular}
}
\caption{\label{tab:disagreed_response_type_examples} Examples of two agent reference responses with different response strategies. 
}
\end{table*}

\begin{table}
\centering
\normalsize
\resizebox{0.49\textwidth}{!}{
\begin{tabular}{p{10cm}}
\toprule
\dots\textbf{<User>:} What kinds of regional varieties are there? \\ 
\textbf{<Agent>:} Would you like to know about East Asia, Southeast Asia, South Asia, or Europe? \\ 
\textbf{<User>:} Tell me about East Asia. \\ \textbf{<Agent>:} Sorry, but each country is detailed as well, do you want to know more about congee in China, Japan, Korea or Taiwan? \\
\bottomrule
\end{tabular}
}
\caption{\label{tab:consecutive_clarifications} An example of consecutive clarifications.}
\end{table}

\paragraph{Agent Response Strategy}
Table~\ref{tab:data_by_rtype} shows the diversity of agent response strategies in \ourdata.
When no direct answer exists, agents in \ourdata can respond to the user with a \textit{relevant answer} (see \S~\ref{sec:annotation}).
If no direct or relevant answer is found, the agent can then respond with \textit{no information}.
The average response length and number of evidence passages differ dramatically across various response strategies.
Compared with direct or relevant answer cases, \textit{clarification} responses tend to be shorter and are more likely to happen when more evidence passages are present.
We also calculate that 30\% \textit{direct} or \textit{relevant answer} agent turns have multiple evidence passages, which potentially require information summarization. 


\subsubsection{Analysis of Agent Initiatives}
\label{sec:agent_initiative}
In this section, we present qualitative analysis to understand how different agent initiatives get triggered, with a focus on \textit{clarification} and \textit{relevant answer} agent responses.
\paragraph{Fine-Grained Categorization} 
We randomly sample and analyze 100 clarification and relevant answer responses respectively. Table~\ref{tab:agent_examples} (upper half) shows that in most cases, the agent raises a clarification when they find a long answer or too many answers (86\%) or notice an ambiguous entity in the user request (13\%). In 70\% of relevant answer cases (bottom half of Table~\ref{tab:agent_examples}), the agent relaxes some constraint in the user request or provides evidence that no definite answer can be found. In 29\% of these cases, they simply provide some relevant but side information only. We also observe that in rare cases (1\%), the agent points out some mistake (e.g., a false assumption) in the user request. 

\paragraph{Clarification Occurrences}
We next look at contexts where
agents are more likely to ask for clarification in a conversation.
Clarification questions occur more frequently at the very beginning (ex.\ 2, Table~\ref{tab:agent_examples}), rather than later in a conversation (18.8\% vs.\ 11.5\%). 
If a clarification is raised in the previous agent turn, the chance of 
a subsequent clarification (see Table~\ref{tab:consecutive_clarifications}) is 7.6\%, compared to 12.2\% if the previous turn is not a clarification (ex.\ 1, Table~\ref{tab:agent_examples}). 

\paragraph{Response Strategy Selection}
In 23\% of examples with 2 agent annotations marked as equally comprehensive by validators, workers take \textit{different response strategies given the same dialogue context}. 
\textcolor{black}{
Of this set, 82\% have different evidence passages labeled by the two workers, potentially due to the open-endedness of user queries in \ourdata and the large knowledge source.
In addition, as suggested by our analyses in \S~\ref{sec:results-analysis}, it is more likely that agents will choose different evidence passages when there is no direct answer to the question.  
As illustrated in the first example in Table~\ref{tab:disagreed_response_type_examples}, 
the different evidence passages often trigger different agent response strategies. 
}
The second and third examples show that even if two agents find the same evidence set, deciding whether it indicates an under-specified user request, a direct or only a relevant answer can be subjective.


\section{Experiment Setup}
\label{sec:exp}

\subsection{Systems}
\label{sec:systems}
We build two systems for each of the tasks formulated in \S~\ref{sec:tasks}. Both systems build on retriever-reader models, inspired by recent advances in open-domain single-turn or conversational question answering \citep{karpukhin-etal-2020-dense, izacard-grave-2021-leveraging, adlakha2022topiocqa}.
Here, the main function of the retriever is to gather a ranked set of top-k candidate evidence passages from the entire Wikipedia to facilitate passage identification and response generation for the later reader model.
We first describe the retriever models, and then introduce the two reader models that perform the two main tasks based on retrieval results. 

\subsubsection{Retriever Models}
\label{sec:ir_models}
We experiment with two retrievers: BM25 and DPR. 
\textbf{BM25} \citep{Robertson2019bm25} uses sparse bag-of-word representations for ranking passages with regard to each query. We use Pyserini \citep{yang-anserini} in our experiments.
\textbf{DPR} \citep{karpukhin-etal-2020-dense} is a BERT-based \citep{devlin-etal-2019-bert} dual encoder model, that produces learned dense representations for queries and passages, and measures the relevance using the dot product similarity in the vector space.
We finetune DPR on \ourdata.
As the training set is small in \ourdata, we initialize it with a downloadable checkpoint\footnote{\url{https://github.com/McGill-NLP/topiocqa}} that is pre-trained on a much larger (> 30$\times$) open-domain conversational question answering dataset, TopioCQA \citep{adlakha2022topiocqa}.

\subsubsection{Reader Models}
\label{sec:reader_models}
Our two readers are based on state-of-the-art models in open-domain question answering and conversational knowledge identification---Fusion-in-Decoder
\citep{izacard-grave-2021-leveraging} and DIALKI \citep{wu-etal-2021-dialki}. 

\paragraph{Fusion-in-Decoder (FiD)} FiD is a generative reader model. It first encodes all retrieved passages with a given query, and then decodes the task output (e.g., an answer string) by attending over all encoded passages. To adapt FiD to our tasks, we prepend a passage identifier (ID) to each of the top-k retrieved passages (here, $k=50$, following \citet{adlakha2022topiocqa})
and separately concatenate each passage with the dialogue context for encoding.
Given the 50 encoded contextualized passage vectors,
the decoder generates a sequence of evidence passage IDs (\textit{passage identification}), followed by the final response (\textit{response generation}). 
After the first turn, the encoded passage vectors associated with $\{\mathcal{P}_i, \dots, \mathcal{P}_{n-1}\}$ are concatenated with the top-k retrieved passages, limiting $k$ to give a total of 50.
In training, we use the same hyperparameters as in \citet{adlakha2022topiocqa}, with the batch size adjusted for the memory constraint and training steps adjusted to have the same epochs. 

\paragraph{DIALKI + FiD} The second reader adopts a pipelined approach to perform the two tasks. It first uses DIALKI \citep{wu-etal-2021-dialki} to select evidence passages and then feeds the identified passages into FiD to generate the agent response. DIALKI is a state-of-the-art conversational knowledge identification model that incorporates dialogue structure with a multi-task learning framework. DIALKI predicts a passage score for each input passage (i.e., each top-k retrieved passage). To adapt it for passage identification, we simply keep evidence passages (up to 4, as in data collection)
with ranking scores higher than $\gamma$ for \textit{multiple passage prediction}, where $\gamma$ is tuned on the dev set. We apply the same method to incorporate previously used evidence passages into DIALKI as in the first reader model. We set the number of input passages of DIALKI to be 50 and keep other original hyperparameters. Parameters in FiD are the same as the first reader model, except that the number of input passages is 4 in the DIALKI+FiD system.

\paragraph{Trivial Baseline: Last Turn} We report performance of a simple baseline: use the most recent agent turn in the dialogue context and associated evidence ($\bar{\mathcal{P}}_{n}=\mathcal{P}_{n-1};\ \bar{a}_n=a_{n-1}$). For first-turn instances, we use the most frequent evidence passage and agent response seen in the training set as the prediction. We also tried using a random previous turn as the prediction, which gives lower scores than using the last turn.

\paragraph{Human} We collect one additional annotation for each agent turn in the test set and evaluate it as the human performance. \textcolor{black}{These additional annotations are annotated by the same agent workers we select in \S~\ref{sec:quality}.} Note that these additional prediction data do not go through the same validation step as those that are used as references.

\subsection{Evaluation}
Below, we describe automatic metrics and a human evaluation protocol for the passage identification (PI) and response generation (RG) tasks in \S~\ref{sec:tasks}.

\paragraph{Passage Identification} \ourdata allows for multiple evidence passages, so we measure the model performance by computing the F1 score (PI-F1), comparing the set of predicted evidence passages $\Bar{\mathcal{P}_n}$ to the set of reference passages  $\mathcal{P}_n$. 
For turns where there are two valid reference annotations, we use the maximum F1 score between the two.

\paragraph{Response Generation} For a generated agent response $\Bar{a}$, we calculate the 
SACREBLEU score \citep{post-2018-call} (BLEU in tables) and token-level F1 (RG-F1) scores against the reference response, following previous studies \citep{feng-etal-2020-doc2dial,adlakha2022topiocqa}. 
Again, when there are two valid annotations, we use the maximum.

\paragraph{Human Evaluation}
As the two tasks are dependent on each other, decoupled automatic evaluations may not capture aspects like factual consistency between predicted passages and the response. Moreover, handling queries with multiple evidence passages or no direct answer can be open-ended.

Therefore, we design a human evaluation protocol to evaluate the model performance on both tasks.\footnote{We release the code at \url{https://github.com/ellenmellon/INSCIT/tree/main/eval/human_eval}.}
Specifically, we focus on the evaluation of 4 dimensions: 1) \textit{evidence passage utility}: how many predicted evidence passages are used in the generated response; 2) \textit{factual consistency} between the predicted response and evidence; 3) response \textit{coherence} with the dialogue context; and 4) response \textit{comprehensiveness}: how much information, that is both relevant to the user request and factually consistent with the predicted evidence, is contained in the response.
While most prior work on information-seeking dialogues only relies on automatic metric scores \citep{choi-etal-2018-quac, anantha-etal-2021-open, adlakha2022topiocqa}, a few studies collect human ratings on dimensions like response ``coherence'' and ``informativeness'' \citep{gao-etal-2022-unigdd,feng-etal-2022-dialdoc}. However, as they do not require models to predict evidence, the factual consistency between the response and the knowledge source cannot be evaluated \citep{nakano2webgpt}.

We provide outputs for both tasks of our two systems and ``Human'' to a human judge. We ask them to rate the first 3 dimensions for each system output on a 4- or 5-point Likert scale\footnote{The 4-point scale is used only for coherence to discourage \textit{neutral} ratings. We report all scores normalized to a 1-5 scale.} and then rank the system responses in terms of \textit{response comprehensiveness} (ties are permitted). 
We have 3 raters for each agent turn and take the average rating score or rank place on each dimension for each system.
Since human evaluation can be time-consuming and costly, we run it on a sampled test subset with 50 conversations (290 turns) and encourage future studies to report on the same subset.


\textcolor{black}{The inter-rater agreement measured as Krippendorf's alpha is 0.66, 0.64, 0.42 and 0.37 for EU, FC, CO and COMP, respectively, which can be interpreted as good or moderate agreements. We observe two main types of coherence disagreements: 1) some workers are more strict and indicate one response as more preferred due to minor differences (e.g., a connecting word), or 2) both responses are incoherent, but in very different ways (e.g., have very different content). Similarly, most comprehensiveness disagreements involve either: 1) two responses that are similar except that one includes additional side information, or 2) two responses that provide different answers but both are good.}

\section{Experiment Results}
\subsection{Quantitative Results}

\begin{table}
\centering
\small
\resizebox{0.8\columnwidth}{!}{
\begin{tabular}{llccccc}\toprule
\textbf{Retriever} & \textbf{Reader} &PI-F1 &BLEU &RG-F1 \\\midrule
Last Turn &Last Turn & 10.5 & 4.2 & 14.1 \\ \midrule
\multirow{2}{*}{BM25} & FiD & 14.1 & 9.4 & 22.5  \\
&DIALKI + FiD & 17.0  & 13.8  & 24.8 \\ \midrule
\multirow{2}{*}{DPR} & FiD & 17.1 & 8.8 & 21.6 \\
&DIALKI + FiD & \textbf{21.5}  & \textbf{16.6} & \textbf{26.6} \\
\bottomrule
\end{tabular}
}
\caption{Automatic scores on the \textbf{dev} set.}
\label{tab:dev_results}
\end{table}


\begin{table}
\centering
\small
\resizebox{\columnwidth}{!}{
\begin{tabular}{llccc|cccc}\toprule
\multirow{2}{*}{\textbf{Retriever}} &\multirow{2}{*}{\textbf{Reader}} &\multicolumn{3}{c}{\textbf{Automatic}} &\multicolumn{3}{c}{\textbf{Human}} \\
&&PI-F1 &BLEU &RG-F1 &EU &FC &CO \\\midrule
\multirow{2}{*}{DPR} & FiD & 17.5 & 9.6 & 22.2 & 2.35 & 2.52 & 3.76 \\
&DIALKI + FiD & \textbf{23.7} & \textbf{16.0} & \textbf{27.8} & \textbf{4.33} & \textbf{4.74} & \textbf{3.77} \\
\midrule
-&Human & 52.5 & 33.8 & 43.5 & 4.76 & 4.77 & 4.85 \\
\bottomrule
\end{tabular}
}
\caption{Automatic scores on the \textbf{test} set, and human scores on 50 \textit{sampled} test conversations (290 turns) for dimensions rated with Likert scales: evidence utility (EU), factual consistency (FC) and coherence (CO).}
\label{tab:test_results}
\end{table}

Table~\ref{tab:dev_results} shows the overall automatic evaluation results of all systems for our main tasks (PI and RG) on the \textit{dev} set. The simple baseline performs very poorly. Using retrieval results from DPR (vs. BM25)  leads to the best overall performance for both tasks. For both BM25 and DPR retrievers, DIALKI + FiD achieves better performance than FiD in all metrics.
A possible reason could be that the smaller number of context vectors used with DIALKI+FiD  is better suited to learning from limited data than the end-to-end FiD approach. 
DIALKI leverages previous evidence passages in passage identification, so its following FiD response generation has only 4 context vectors (vs.\ 50 for FiD).
This hypothesis is supported by the observation that incorporating previously used evidence hurts the RG performance slightly for FiD but for DIALKI+FID it helps (roughly 1 point decrease vs.\ increase in scores, respectively, with DPR).

\begin{figure}
\centering
\includegraphics[width=1.0\linewidth]{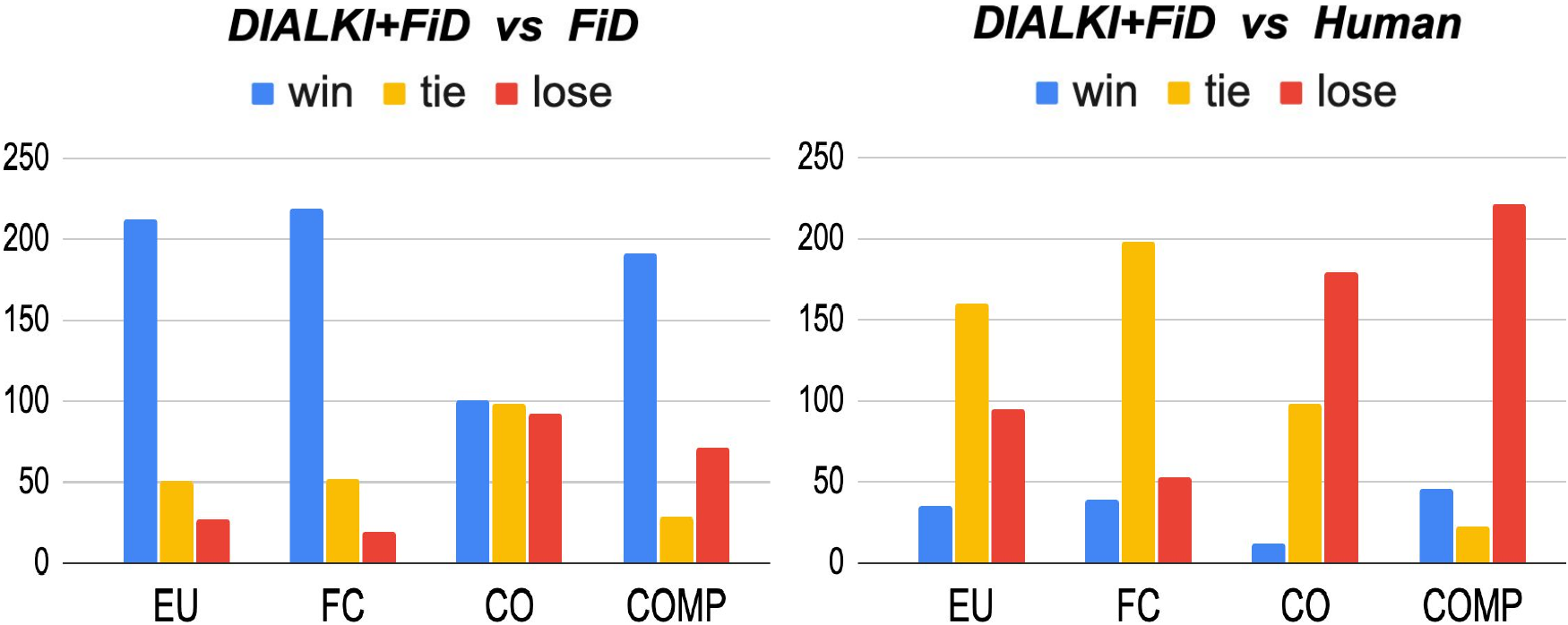}
\caption{\label{tab:human_eval_compare} Human evaluation on system comparison for all dimensions: evidence utility (EU), factual consistency (FC), coherence (CO) and response comprehensiveness (COMP). Win/lose refers to DIALKI+FiD. }
\end{figure}

Table~\ref{tab:test_results} shows both automatic and human evaluation results on the \textit{test} set for FiD and DIALKI+FiD with the DPR retriever, confirming the dev set findings.
Experiments with BM25 also confirm dev set trends.
Figure \ref{tab:human_eval_compare} presents comparative human evaluation results.
DIALKI+FiD greatly outperforms FiD except in coherence where scores are similar. 
DIALKI+FiD substantially underperforms humans in both automatic and human scores, except for factual consistency where 
the difference is small. This could indicate that, although DIALKI+FiD generates responses consistent with the predicted evidence, it identifies less relevant passages which lead to less coherent and less informative responses.

The reason for \textit{imperfect human performance} on passage identification, shown in Table~\ref{tab:test_results}, is two-fold. Due to the
open-endedness of information-seeking queries in \ourdata and the large search space over Wikipedia, annotators may find different (but both valid) sets of evidence passages. In addition, annotations corresponding to the Human ``system'' do not go through the validation process, so they could have errors or be less comprehensive.

\subsection{Analysis}
\label{sec:results-analysis}
\paragraph{Passage Retrieval} Table~\ref{tab:ir_performance} reports the performance for passage retrieval in HIT@k scores, following \citet{karpukhin-etal-2020-dense,adlakha2022topiocqa}. HIT@k is calculated as $\mathbbm{1} \big[|\mathcal{R}_K \cap \,\mathcal{P}| > 0\big]$, where $\mathcal{R}_K$ denotes the top $K$ retrieved passages and $\mathcal{P}$ 
denotes the union of the two reference passages sets (or a single reference set if only one is valid).
We evaluate both BM25 and DPR models used in our main experiments, as well as two DPR ablations: with pretraining (PT) on TopioCQA or finetuning (FT) on \ourdata only. BM25 underperforms DPR models significantly, which explains the main task performance differences between BM25 and DPR in Table~\ref{tab:dev_results}. DPR with PT alone is more effective than FT only, which can be explained by the much larger
training data in TopioCQA. 
The best retrieval results are achieved with PT and FT combined.
We do not leverage TopioCQA for pretraining on our two main tasks, because 1) it does not come with the passage identification task and only has short answers or \texttt{no answer} as their agent responses; 2) we observe poor zero-shot response generation performance on \ourdata for FiD trained on TopioCQA. 


\begin{table}
\centering
\resizebox{0.8\columnwidth}{!}{
\begin{tabular}{lcc|ccc}\toprule
\multirow{2}{*}{\textbf{Retriever}} &\multicolumn{2}{c}{\textbf{Dev}} &\multicolumn{2}{c}{\textbf{Test}} \\
 & HIT@20 & HIT@50 & HIT@20 & HIT@50\\ \midrule
BM25 & 35.3 & 48.0 & 35.6 & 48.1 \\
DPR (FT only) & 62.5 & 70.1 & 51.3 & 60.8 \\
DPR (PT only) & 66.4 & 76.3 & 68.4 & \textbf{77.5} \\
DPR & \textbf{71.1} & \textbf{79.8} & \textbf{69.9} & \textbf{77.5} \\
\bottomrule
\end{tabular}
}
\caption{\textit{Passage retrieval} results. PT and FT refer to pretraining on TopioCQA and finetuning on \ourdata.}
\label{tab:ir_performance}
\end{table}

\begin{figure}
\centering
\includegraphics[width=1.0\linewidth]{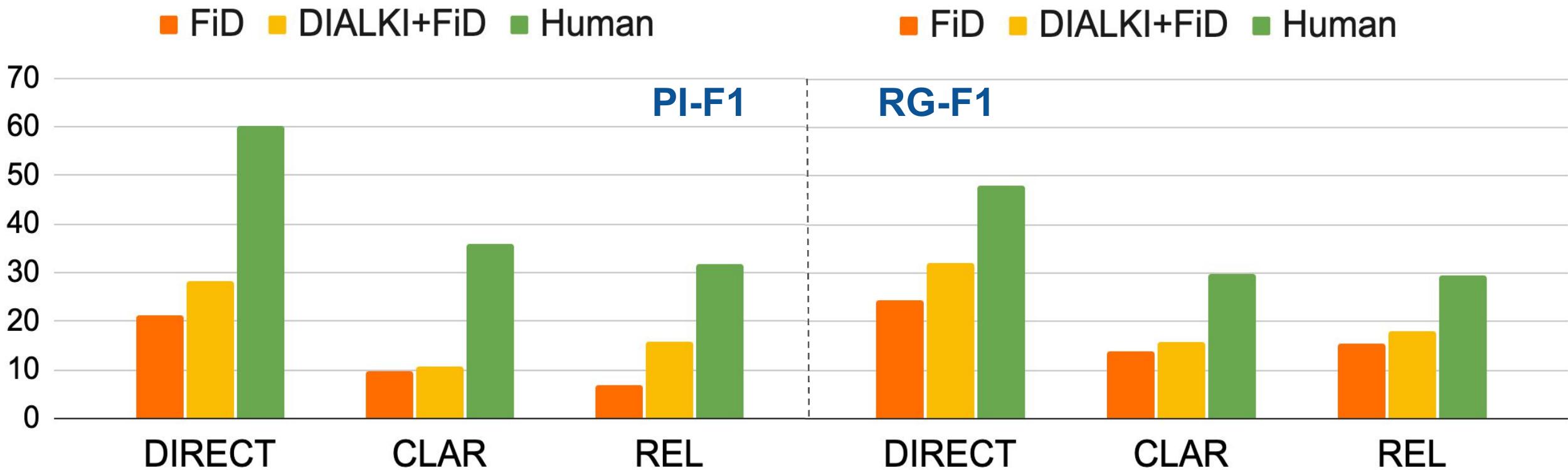}
\caption{\label{fig:resp-strategy2} PI-F1 and RG-F1 scores by reference response strategy (direct answer, clarification, relevant answer) on the one-strategy test subset, excluding instances where two references differed in strategy.}
\end{figure}

\paragraph{Passage Identification \& Response Generation Performance Breakdown}
Figure~\ref{fig:resp-strategy2} shows the system and task performance breakdown by \textit{reference response strategy} (direct answer, clarification and relevant answer) for the test set, excluding examples where two annotations differed in the response strategy category (16\%). DPR is used for retrieval.
Only RG-F1 is shown for response generation; trends for BLEU are similar. For all response types, DIALKI+FiD is similar or outperforms FiD, but significantly underperforms humans. For both systems and humans, the non-direct-answer responses have lower automatic scores. The lower PI-F1 scores for humans suggest that the retrieval task is more difficult (with more variety in evidence) when a simple direct answer is not available. Lower automatic response generation scores may be explained by lower retrieval scores (less reliable evidence), larger number of passages, and/or challenges in learning non-direct-answer response strategies.  Note that for both systems, the largest percentage gap with respect to human scores is for clarifications. 


\paragraph{Response Generation Results with Human Evidence Passages} To explore the above hypotheses, we generated responses using the DIALKI+FiD response generator with passages selected in the ``Human'' annotation of the test data. The resulting responses had 26.5 and 37.4 for BLEU and RG-F1 scores, respectively, compared to 16.0 and 27.8 when using DIALKI passages. 
We sample and analyze 20 examples each of single and multiple ``Human'' evidence passages. Given multiple evidence passages, most DIALKI+FiD responses either do not use all passages or introduce incorrect facts. With one passage, responses are consistent with the evidence but not always as comprehensive as for humans. In the 20 examples with multiple passages, DIALKI+FiD asks one clarification, whereas humans ask nine.

\textcolor{black}{
\paragraph{Impact of Response Type Prediction for Response Generation} 
As explained in \S~\ref{sec:analysis}, the agent response type depends on the selected evidence passages. 
To analyze how incorporating response types can help with response generation, we conduct a controlled experiment to generate agent responses with the dialogue context and oracle evidence passages 
as the input to FiD, and compare the performance when no/oracle/predicted response type is given. For examples that have two labels with different sets of evidence passages, we split them into two separate instances. 
To predict the response type, we use a sequence classification model based on BERT-base \citep{devlin-etal-2019-bert}, given the dialogue context and oracle evidence passages. To provide the oracle or predicted response type as the response generation model input, we simply append a formatted string---\texttt{response type: \{response\_type\_name\}}\footnote{Candidate response type names are ``direct answer,'' ``clarification,'' ``no answer but relevant information'' and ``no answer and no information.''}---at the end of the dialogue context, when feeding it to FiD.} 

\textcolor{black}{The response type classification model gives an overall accuracy of 0.75, compared to 0.73 when  predicting everything as ``direct answer.''
Table~\ref{tab:rg_by_rtype} shows that adding either oracle or predicted response types improves BLEU and RG-F1 scores, compared with no response type being used, with greater gains in RG-F1 for oracle response type. 
We observe consistent performance gains on examples with either ``direct answer'', ``clarification'' or ``no information'' oracle response types, but not for the ``relevant answer'' response type.
}

\begin{table}
\centering
\resizebox{0.99\columnwidth}{!}{
\begin{tabular}{lcc|cc}\toprule
\multirow{2}{*}{\textbf{Model Input}} & \multicolumn{2}{c}{\textbf{Dev}} & \multicolumn{2}{c}{\textbf{Test}} \\ 
& BLEU & RG-F1 & BLEU & RG-F1 \\\midrule
DC+OEP+RT (Oracle) & 32.6 & 48.7 & 31.6 & 47.4  \\
DC+OEP+RT (Predicted) & 32.6 & 46.3 & 31.7 & 45.4 \\
DC+OEP & 32.0 & 45.3 & 30.6 & 44.3 \\
\bottomrule
\end{tabular}
}
\caption{\textcolor{black}{Automatic RG scores for FiD with inputs: dialogue context (DC), oracle evidence passages (OEP), and different (oracle/predicted/no) response types (RT).}}
\label{tab:rg_by_rtype}
\end{table}

\section{Conclusion \& Discussions}

In summary, we introduce \ourdata, a new open-domain information-seeking conversational dataset grounded in Wikipedia, with mixed-initiative user-agent interactions. \ourdata supports two tasks (passage identification and response generation), for which we present results of two strong baselines, with best results obtained with the pipelined DIALKI+FiD system.
We also introduce a human evaluation protocol.

\paragraph{Future Work}
Both models significantly underperform humans in both tasks in all metrics. The relative performance gap is greatest for scenarios that require the agent to provide a non-direct answer.
We find that passage identification significantly impacts response generation (particularly coherence) by providing relevant grounding knowledge. 
Thus, improving methods for selecting relevant passages is critical for future work.
Key challenges that remain in response generation are how to fuse and present comprehensive information from multiple passages and learning when and how to use
non-direct response strategies. 
Given the small size of our training data, another future direction is to explore transfer learning using existing information-seeking conversation or question answering resources.

\textcolor{black}{
Our work focuses on different strategies that can be adopted by the agent to better address user requests in a conversational question answering setting, assuming the user will either ask an information-seeking question or provide a clarification to the agent. 
Exploring more user-side strategies would be interesting for handling system errors and other types of conversations (e.g. negotiations).
}

\textcolor{black}{In contrast to Wikipedia passages, information sources used in practice (e.g., the whole web) can often contain less trustworthy information. In such cases, retrieving evidence passages containing the same answer and predicting the trustworthiness of each answer based on all such retrieved passages can be a promising direction.}

\textcolor{black}{Another direction that is worth future exploration lies in the design of evaluation metrics. We follow previous studies to evaluate the model performance when given a fixed human-human dialogue context. However, as pointed out by \citet{li-etal-2022-ditch}, an interactive dialogue system often needs to handle dialogue contexts containing errors made by the model itself. Therefore, 
it is important for future work to develop new methods for automatic evaluation and scalable human evaluation in the interactive setting.}

\section{Ethical Considerations for Dataset Collection}
Our work is primarily intended to encourage future work in information-seeking conversation \textit{factually grounded} in given knowledge sources. Our knowledge sources come from Wikipedia articles, where the content follows principles emphasizing on a neutral point of view and reliable sources. Before and during the data collection, we carefully guide our user workers not to ask subjective or opinion-driven questions, and our agent workers not to include any content without evidence from the knowledge sources in their conversational responses. Therefore, all contents exposed to our workers during data collection should contain minimal risk to the workers. Our data collection has IRB approval from University of Washington and is deemed exempt. We also actively communicated with the workers to address any concern they had and we usually replied back within an hour during the whole data collection process. This communication also helped us to make sure that our workers were compensated fairly. As explained in \S~\ref{sec:quality}, most of our workers report that they
are paid with an hourly rate of 15-20 USD.

\section*{Acknowledgments}
This research was supported by NSF IIS-2044660, ONR N00014-18-1-2826, DARPA MCS program  (N66001-19-2- 4031), a Sloan fellowship and gifts from AI2.
We thank Tao Yu and all members in H2Lab and TIAL labs at University of Washington who participated in our data collection pilot studies. We thank Kevin Everson for proofreading our paper and providing valuable suggestions.
We also thank members of University of Washington's NLP groups who provided feedback and insights to this work.
We specifically want to thank our TACL reviewers and action editor for their detailed comments that lead to great improvement of our paper.



\bibliography{custom}
\bibliographystyle{acl_natbib}

\end{document}